\documentclass[10pt,twocolumn,letterpaper]{article}

\usepackage{cvpr}
\usepackage{times}
\usepackage{epsfig}
\usepackage{graphicx}
\usepackage{amsmath}
\usepackage{amssymb}
\usepackage{multirow}
\usepackage{color}

\usepackage[pagebackref=true,breaklinks=true,letterpaper=true,colorlinks,bookmarks=false]{hyperref}

 \cvprfinalcopy 

\ifcvprfinal\fi
\begin{document}

\title{3D Face Reconstruction from Light Field Images: A Model-free Approach \\ (\textcolor{blue}{\textbf{This a preprint of the paper accepted in ECCV 2018}}) }


\author{Mingtao Feng $^{1,2}${\hspace{4mm}} Syed Zulqarnain Gilani$^{1}$ {\hspace{4mm}} Yaonan Wang $^2$ {\hspace{4mm}}  Ajmal Mian$^{1}$\\
$^1$ School of Computer Science and Software Engineering,\\
The University of Western Australia\\
{\tt\small \{zulqarnain.gilani,ajmal.mian\}@uwa.edu.au}
\and
\vspace{-3mm}\\
$^2$College of Electrical and Information Engineering, \\
Hunan University\\
{\tt\small \{mintfeng,yaonan\}@hnu.edu.cn}
}

\maketitle

\begin{abstract}
Reconstructing 3D facial geometry from a single RGB image has recently instigated wide research interest. However, it is still an ill-posed problem and most methods rely on prior models hence undermining the accuracy of the recovered 3D faces. In this paper, we exploit the Epipolar Plane Images (EPI) obtained from light field cameras and learn CNN models that recover horizontal and vertical 3D facial curves from the respective horizontal and vertical EPIs. Our 3D face reconstruction network (FaceLFnet)  comprises a densely connected architecture to learn accurate 3D facial curves from low resolution EPIs.  To train the proposed FaceLFnets from scratch, we synthesize photo-realistic light field images from 3D facial scans. The curve by curve 3D face estimation approach allows the networks to learn from only 14K images of $80$ identities, which still comprises over 11 Million EPIs/curves. The estimated facial curves are merged into a single pointcloud to which a surface is fitted to get the final 3D face. Our method is model-free, requires only a few training samples to learn FaceLFnet and can reconstruct 3D faces with high accuracy from single light field images under varying poses, expressions and lighting conditions. Comparison on the BU-3DFE and BU-4DFE datasets show that our method reduces reconstruction errors by over 20\% compared to recent state of the art.

\end{abstract}

\vspace{-5mm}
\section{Introduction}

\begin{figure}  
	\center{\includegraphics[width=0.5\textwidth]{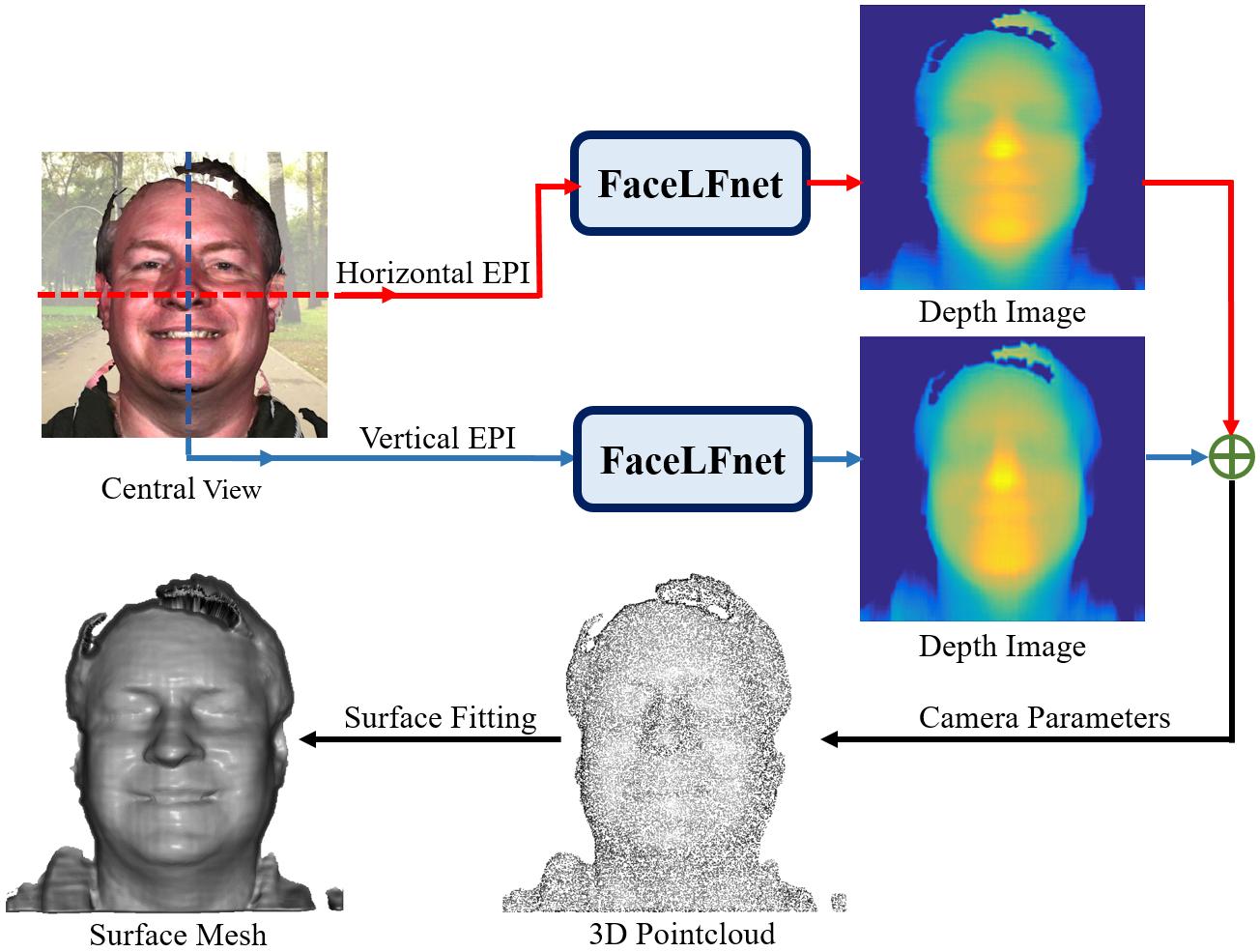}}
	\caption{\label{Overview} Proposed pipeline for 3D face reconstruction from a single light field image. Using synthetic light field face images, we train two FaceLFnets for regressing 3D facial curves over their respective horizontal and vertical EPIs. The estimated depth maps are combined, using camera parameters, into a single pointcloud to which a surface is fitted to get the final 3D face.}
	\vspace{-2mm}
\end{figure}

Three dimensional  face analysis has the potential to address the challenges  that confound its two dimensional counterpart such as variations in illumination, pose  and scale~\cite{abate2007}. This modality has achieved state-of-the-art performances on applications such as face recognition~\cite{gilani2017b,Ajmal2007_PAMI,queirolo2010}, syndrome diagnosis~\cite{Fchild2007,peter2008,tan2017,whitehouse2015}, gender classification~\cite{gilani2014c} and face animation~\cite{Fanmation2013_ACM,Fanmation2016_CVPR}. Reconstructing 3D facial geometry from RGB images is, therefore, receiving a significant interest from the research community. However, using a single RGB image to recover the 3D face is an ill-posed problem \cite{Reference3D2011_PAMI} since the depth information is lost during the projection process. In fact, many different 3D shapes can result in similar 2D projections. The scale and bas-relief ambiguities \cite{Belhumeur1999} are common examples.

Most existing methods have resorted to the use of prior models such as the Basal Face Model (BFM)~\cite{bfm} and the Annotated Face Model(AFM) \cite{afm} to generate synthetic data with ground truth to train CNN \cite{E2E3Dface2017_CVPR,3DVface2016} models and to recover the model parameters at test time. However, model-based approaches are inherently biased and constrained to the space of the training data of the prior models.

A 4D light field image captures the RGB color intensities at each pixel as well as the direction of incoming light rays. High resolution plenoptic cameras~\cite{Lytro,Raytrix} are now commercially available. Plenoptic cameras use an array of micro-lenses to capture many sub-aperture images arranged in an equally spaced rectangular grid. Unlike most 3D scanners that use active light projection and are hence restricted to indoor use, plenoptic cameras are passive and can instantly acquire light field images outdoors as well, in a single photographic exposure. The sub-aperture light field images have been exploited to improve the performance of many applications such as saliency detection~\cite{Yu2015weighted}, hyperspectral light field imaging~\cite{Yu2017_CVPR}, material classification~\cite{Wang2016material}, image segmentation~\cite{Zhu_seg2017_CVPR} and image restoration~\cite{Tian_Restor2017_ICCV,Wu_Recons2017_CVPR} and in particular, depth estimation~\cite{Jeon2015depth,Tao2015depth,Yu2015depth,Wang2015occlusion,Sheng2017occlusion}. This paper focuses on reconstructing 3D faces from light field images under a wide range of pose, expression and illumination variations.

Various methods have been proposed to solve the ill-posed problem of reconstructing 3D facial geometry from a single RGB image~\cite{Reference3D2011_PAMI,3DVface2016,E2E3Dface2017_CVPR,RDverydeep2017_CVPR,FacialI2I2017_ICCV,Fpose2017IJCV}. These methods all use one or more common techniques. For instance, Shape from Shading (SfS) uses the shading variation to reconstruct 3D faces but the caveat is that the method is sensitive to lighting and RGB image texture and even under near ideal conditions, suffers from the bas-relief ambiguity \cite{Belhumeur1999}. 3D Morphable Models (3DMM)~\cite{E2E3Dface2017_CVPR,3DVface2016}  project the 3D faces in a low-dimensional subspace. However, the models are confined to the linear space of their training data and do not generalize well to all face shapes~\cite{gilani2017a}. Landmark based methods use facial keypoints to guide the reconstruction process but rely heavily on accurate localization of the landmarks.

We propose a model-free approach (see Fig.~\ref{Overview}) to reconstruct 3D faces directly from light field images using Convolutional Neural Networks (CNN). Our technique does not rely on model fitting or landmark detection. Training a CNN requires massive amount of photo-realistic labeled data. However, there is no publicly available 4D light field face dataset with corresponding ground truth 3D face models. We address this problem and propose a method of generating the training data. We use the BU-3DFE~\cite{BU3D_face2006} and BU-4DFE datasets~\cite{BU4D2013} to generate light field images from their ground truth 3D models. Figure \ref{Syn_data} shows some examples. We randomly vary the light intensity and pose to make our dataset more realistic. Our dataset comprises approximately 19K photo-realistic light field images with ground truth depth maps~\footnote{We use depth map to represent disparity map as they are related by light field camera parameters~\cite{4DLFdataset2016}.}. Furthermore, we show that our method requires fewer training samples (facial identities) as it capitalizes on reconstructing 3D facial curves rather than the complete face at once. We believe that our synthesized dataset of 4D light field images with corresponding 3D facial scans can be applied to many other facial analysis problems such as pose estimation, recognition and alignment.

Equipped with a rich light field image dataset, we propose a densely connected CNN architecture (FaceLFnet) to learn 3D facial curves from Epipolar Plane Images (EPIs). We train two networks separately using horizontal and vertical EPIs to increase the accuracy of depth estimation. The densenet architecture is preferred as it can accurately learn the subtle slopes in low resolution EPIs\footnote{Higher slope of lines in EPI corresponds to lower depth values.}. FaceLFnets are trained using our synthetic light field face images for which the ground truth depth data is available. Once the face curve estimates are obtained independently from the horizontal and vertical FaceLFnets, we merge them into a single pointcloud based on the camera parameters and then use a surface fitting method to recover the final 3D face. The core idea of our work is a model-free approach, where the solution is not restricted to any statistical face space. This is possible by exploiting the shape information present in the Epipolar Plane Images.

Our contribution are:
(1) A model-free approach for 3D face reconstruction from a single light field image. Our method does not require face alignment or landmark detection and is robust to facial expressions, pose and illumination variations. Being model-free, our method also estimates the peripheral regions of the face such as hair and neck.
(2) A training technique that does not require massive number of facial identities. Exploiting the EPIs, we demonstrate that the proposed FaceLFnet can learn from only a few identities ($80$) and still outperform the state-of-the-art methods by a margin of $26\%$.
(3) A data syntheses technique for generating a light field face image dataset which, to the best of our knowledge, is the first of its kind. This dataset will contribute to solving other face analysis problems as well.

\vspace{-1mm}
\section{Related Work}
\vspace{-1mm}
3D face reconstruction from a single image has attracted significant attention recently. Shape-from-shading (SfS) has been a popular approach for this task~\cite{SYsfs2000_CVPR,RGBD_SfS2015_CVPR,SfS2013_ICCV}. For example, WenYi \etal~\cite{SYsfs2000_CVPR} proposed a symmetric SfS method to obtain illumination-normalized image and  developed a face recognition system. Roy \etal~\cite{RGBD_SfS2015_CVPR} proposed an improved SfS method to enhance the depth map combining the RGB image and rough depth image to create more details. Yudeog \etal~\cite{SfS2013_ICCV} estimated lighting variations with both global and local light models. SfS approach was then applied with the estimated lighting models for accurate shape reconstruction. Reconstruction using SfS requires priors of reflectance properties and lighting conditions and suffers from the bas-relief ambiguity \cite{Belhumeur1999}.

A 3D Morphable Model (3DMM) was introduced by Blanz and Vetter~\cite{3DMM1999} which represents a 3D face as a linear combination of orthogonal basis vectors obtained by PCA over 100 male and 100 female identities. James \etal~\cite{3DMM_2017_CVPR} extended the concept and proposed a statistical model combined with a texture model for fitting the 3DMM on face images in \textit{the wild}. 3DMM has also been used in~\cite{3DMM2009_CVPR,3DMM2013PAMI,3DMM2016_CVPR,3DMM2014_3DV} for face reconstruction. The main limitation of such methods is that the 3DMM cannot model every possible face. Moreover, it is unable to extract facial details like wrinkles and folds because such details are not encoded in the linear subspace.

Recently, various attempts were made to integrate 3DMMs with CNN for facial geometry reconstruction from a single image. Elad \etal~\cite{3DVface2016} employed an iterative CNN trained with synthetic data to estimate 3DMM vectors. The predicted geometry was then refined by the real-time shape-from-shading method. Matan \etal~\cite{DetailedFace2017_CVPR} extended the work~\cite{3DVface2016} and introduced an end-to-end CNN framework that recovers the coarse facial shape using a \textit{CoarseNet}, followed by a \textit{FineNet} to refine the facial details. The two net parts are connected by a novel layer that renders the depth image from 3D mesh. Pengfei Dou \etal~\cite{E2E3Dface2017_CVPR} proposed an end-to-end 3D face reconstruction method from a single RGB image. They trained a fusion-CNN with multi-task learning loss to simplify 3D face reconstruction into neutral and expressive 3D facial parameters estimation. Jourabloo \etal~\cite{Fpose2017IJCV} proposed a 3DMM fitting method for face alignment, which uses a cascaded CNN to regress camera matrix and 3DMM parameters. Tuan Tran \etal~\cite{RDverydeep2017_CVPR} used multi image 3DMM estimates as ground truth and then trained a CNN to regress 3DMM shape and texture parameters from an input image.

Kemelmacher el at.~\cite{Reference3D2011_PAMI} used the input image as a guide to build a single reference model to align with the face image and then refined the reference model using SfS method. Tal \etal~\cite{Ffrontal2015_CVPR} used a 3D neutral face as reference model to approximate the RGB image for face frontalization. Matan \etal~\cite{FacialI2I2017_ICCV} proposed a translation network that learns two maps (a depth image and a correspondence map), used for non-rigid registration with a template face, from a single RGB image. Fine-tuning is then performed for reconstructing facial details. In contrast to SfS and model fitting based face reconstruction methods, we learn 3D face curves from EPIs of the light field image. Our method does not require face alignment, dense correspondence or model fitting steps and is robust to facial pose, expressions and illumination. 

To the best of our knowledge none of the existing methods is model-free and uses a prior face model at some stage of the reconstruction process. On the other hand our method is completely model-free. Similarly, we are unaware of any existing technique that uses light field images for 3D face reconstruction. However, literature points to some research in shape reconstruction from light field images using deep learning. Heber \etal~\cite{Heber2016_CVPR} presented a method for reconstructing the shape from light field images that applies a CNN for pixel wise depth estimation from EPI patches. Although this method produces accurate scene depth, it uses a carefully designed dataset containing drastic slope changes in the EPIs. This method is unsuitable for non-rigid facial geometry reconstruction as faces are generally smooth and their EPIs contain only subtle slope variations. Heber \etal~\cite{Heber2016BMVC} proposed a U-shaped network architecture that automatically learns from EPIs to reconstruct their corresponding disparity images. However, training the network requires disparity maps of all the light field sub-views as labels, which is unrealistic for real datasets. Our approach differs in three ways. Firstly, we use one full EPI as input and its corresponding depth values as labels to overcome the problem of inaccurate depth estimation in the presence of subtle slope variations in the EPIs. Secondly, we train networks using horizontal EPIs and vertical EPIs separately to obtain a more accurate combined 3D pointcloud. Finally, our method does not require disparity maps of all the light field sub-views.

\vspace{-1mm}
\section{Facial Light Field Image Dataset Generation}
\vspace{-1mm}

\begin{figure}  
  \center{\includegraphics[width=0.5\textwidth]{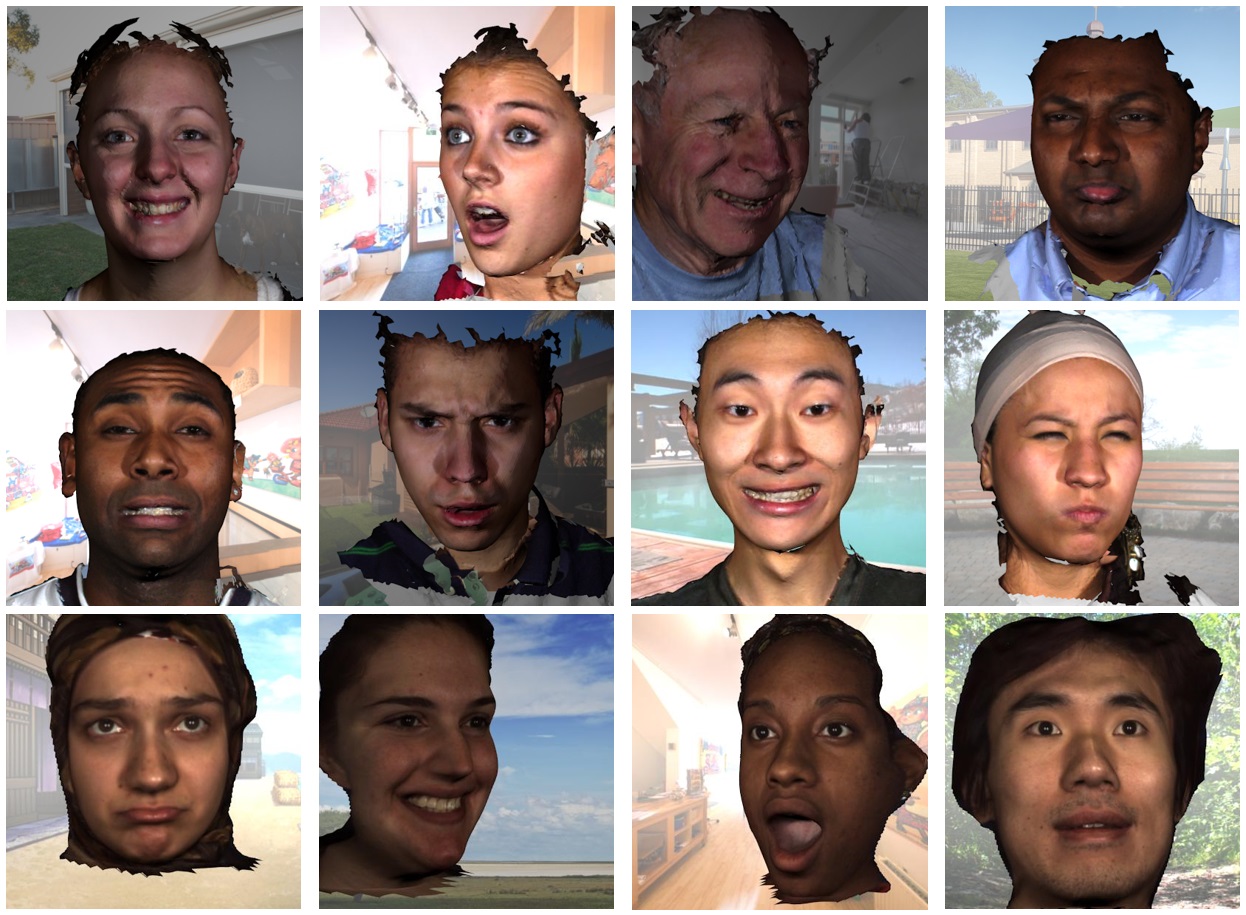}}
  \caption{\label{Syn_data} Central view examples of our rendered light field images. The ground truth 3D scans are aligned with the central view. To make the dataset rich in variations, the generated light field images use random backgrounds and differ extensively in ethnicity, gender, age, pose and illumination.}
  \vspace{-2mm}
\end{figure}

The key to the success of CNN-based 3D face reconstruction from a single RGB image lies in the availability of large training datasets. However, there is no large scale dataset available that provides RGB face images and their corresponding high quality 3D models. 
Similarly, training a light field face reconstruction network requires a large-scale light field face dataset with corresponding ground truth 3D facial scans.  Over the past few years, the computer vision community has made considerable efforts to collect light field images~\cite{4DLFdataset2016,Wang2016material,Yu2014_CVPR,MIT_LFdata,Stanford_LFdata} for different applications. The only public light field face dataset~\cite{IETfacedata2017} captured by Lytro Illum\textsuperscript{TM} camera consists of $100$ identities with $20$ samples per person. However, depth maps of this dataset are generated using the Lytro Desktop Software\textsuperscript{TM}  and have low resolution as well as low depth accuracy. Therefore, this dataset is not suitable for training a network.

\begin{figure}[t]  
	\center{\includegraphics[width=0.45\textwidth]{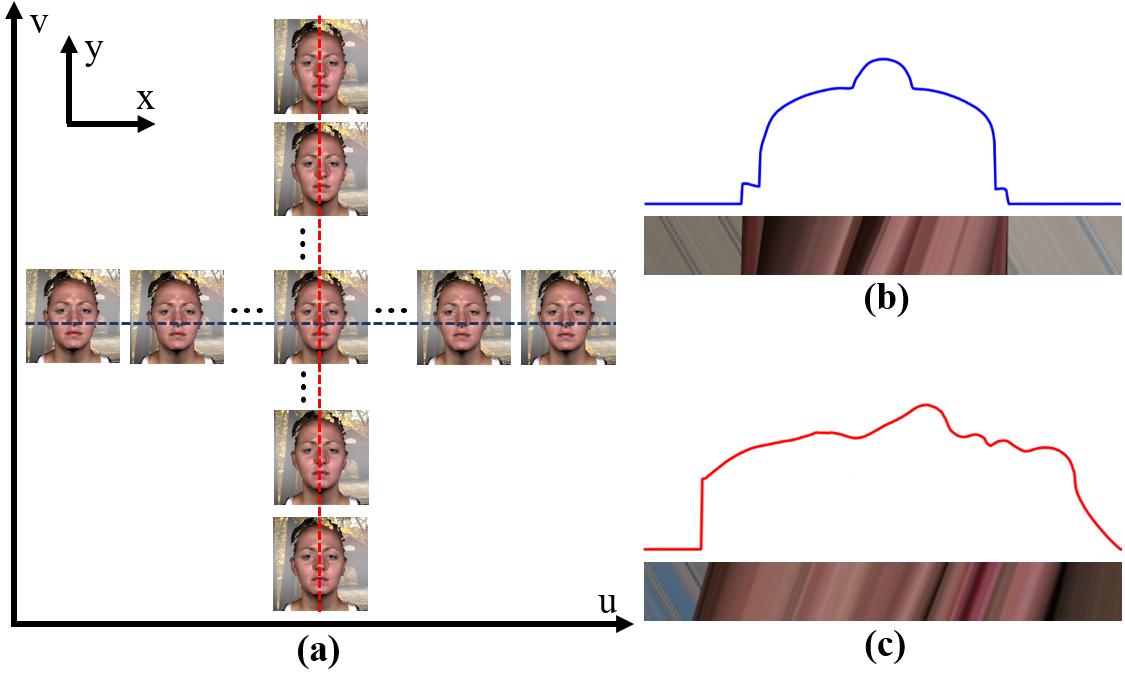}}
	\caption{\label{EPI_faceline} EPIs corresponding to the 3D face curves. (a) Horizontal and vertical EPIs are obtained between the central view and sub-aperture images that are in the same row and column. (b) and (c) Visualization of the relationship between depth curves and slopes of lines in horizontal and vertical EPIs respectively.}
\vspace{-2mm}
\end{figure}

In the absence of large-scale 4D light field face datasets, we propose to generate a dataset of light field face images with ground truth 3D models. For this purpose, we use the public BU-3DFE~\cite{BU3D_face2006} and BU-4DFE~\cite{BU4D2013} databases to generate light field face images. The former is used for training and testing whereas the latter is used only for testing only. The BU-3DFE dataset consists of 2,500 3D scans from $100$ identities ($56\%$ female, $44\%$ male), with an age range from 18 to 70 years and multiple ethnicities. Each subject is scanned in one neutral and $6$ non-natural expressions each with four intensity levels. The BU-4DFE dataset contains 3D video sequences of 101 identities ($58$ female and $43$ male) in six different facial expressions.  We select the most representative frame of each expression sequence. As a result, our dataset contains  $606$ 3D scans. These models contain shape details such as wrinkles of not only the fiducial area, but also the hair, ears and neck area which pose challenges for conventional 3D face reconstruction methods. All 3D models have RGB texture.

To generate plausible synthetic light field face images, it is crucial to control the light field camera parameters, background and illumination properly during rendering. We use the open source Blender\footnote{http://www.blender.org} software and the light field camera tool proposed by Katrin Honauer \etal~\cite{4DLFdataset2016} for this purpose. We place a virtual light field camera in Blender with $15\times15$ micro-lenses and set its field of view to capture the 3D facial scans. Both BU-3DFE and BU-4DFE databases provide 3D facial models in the near frontal pose. We load the 3D models along with their textures in Blender and apply two rigid rotations ($\pm15^\circ$) in pitch and four in yaw ($\pm15^\circ$ and $\pm30^\circ$). To synthesize photo realistic light field images, we apply randomly selected indoor and outdoor images as backgrounds. We place two lamps at different locations in the scene and randomly change their intensities to achieve lighting variations. The angular resolution of the synthetic light field image is $15\times15$ and the spatial resolution is $400\times400$. The ground truth depth maps are aligned with the central view of light field image. Examples of our synthetic light field images are shown in Figure~\ref{Syn_data}.

We implement a Python script\footnote{The script for light field facial image synthesis will be made public.} in Blender on a 3.4 GHz machine with 8GB RAM to automatically generate the light field facial images. The process of synthesizing light field images can be parallelized since each sub-aperture image is rendered independently. In total, we use $80$ identities from BU-3DFE dataset to synthesize 14,000 light field images with ground truth disparity maps. The remaining $20$ subjects from BU-3DFE and all $101$ subjects from the BU-4DFE dataset are used as test data to generate 1,451 light field facial images for evaluation.

\begin{figure}[t]  
	\center{\includegraphics[width=0.45\textwidth]{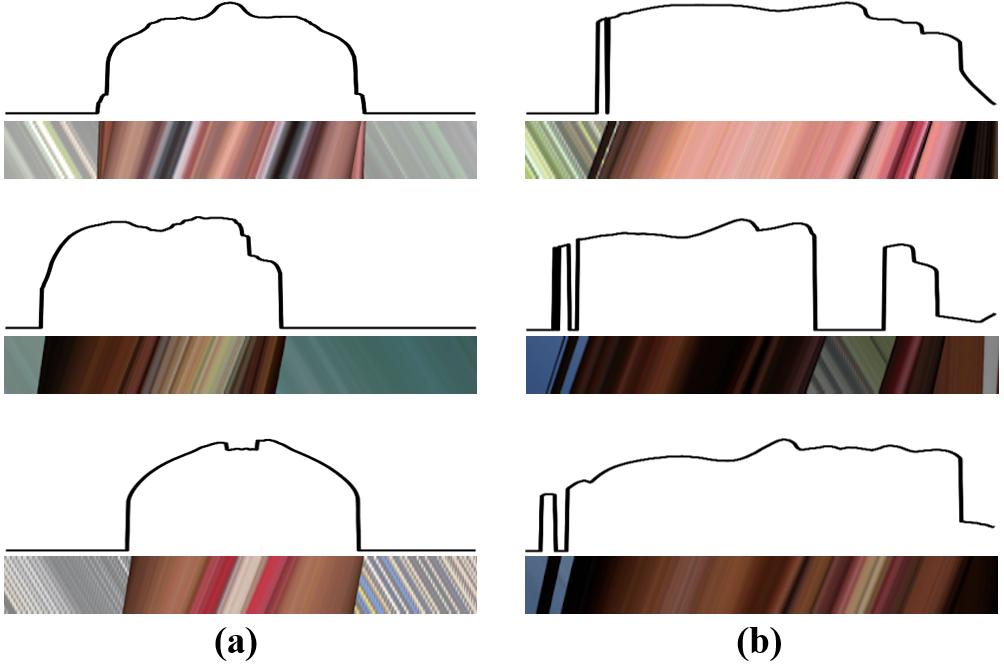}}
	\caption{\label{EPI_example} Examples of EPIs and their corresponding 3D face curves.(a) Horizontal EPIs. (b) Vertical EPIs.}
\vspace{-2mm}
\end{figure}

\vspace{-1mm}
\section{Proposed Method}
\vspace{-1mm}
An overview of the proposed method for reconstructing facial geometry from light field image is shown in Figure~\ref{Overview} and the details follow.

\begin{figure*}[t] 
	\center{\includegraphics[width=\textwidth]{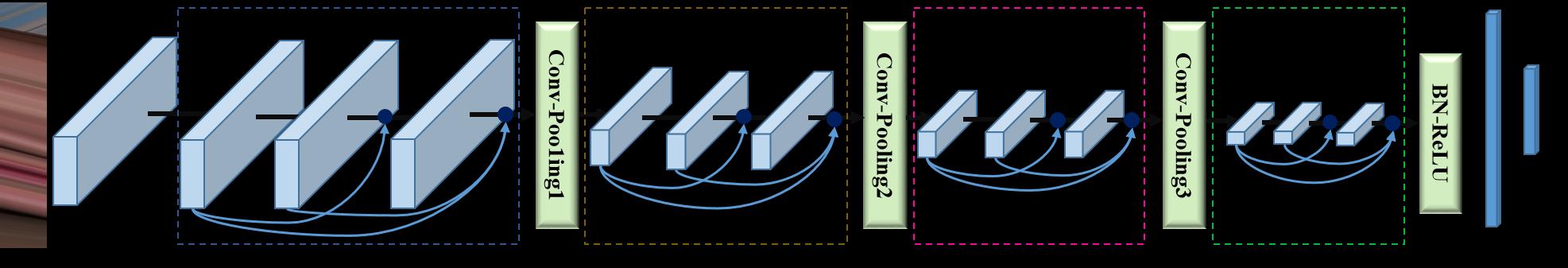}}
	\caption{\label{LFface_CNN} Our proposed FaceLFnet for learning 3D face curves from EPIs. It contains 4 dense blocks, followed by two fully connected layers. The layers between two neighboring blocks are defined as transition layers and change feature map sizes via convolution and pooling~\cite{DenseNet2017_CVPR}. }
\vspace{-2mm}
\end{figure*}

\subsection{Training Data}
\vspace{-1mm}
A 4D light field image can be parameterized as $L(u,v,x,y)$, where $(x,y)$ and $(u,v)$ represent the spatial and angular coordinates respectively~\cite{Wanner2014_PAMI}. When we fix $v$ and $y$, then $L(u,v^*,x,y^*)$ defines a 2D horizontal EPI. Similarly the 2D vertical EPI can be represented as $L(u^*,v,x^*,y)$ when we keep $u$ and $x$ constant. As shown in Figure~\ref{EPI_faceline}, 2D EPIs demonstrate the linear characteristic of the light field image. The orientations of lines within the EPIs can infer the disparity of the corresponding 3D space points~\cite{Wanner2014_PAMI,Sparse2016_CVPR,SPO2016_CVIU,Heber2016_CVPR,Heber2016BMVC}. Equation~(\ref{slopefunction1}) shows the relationship between the slope of the line and the disparity value where $f$ is the light field camera parameter and $k$ is the slope of the line.
\begin{equation}\label{slopefunction1}
Z=-f\times k,
\end{equation}
As shown in Figure~\ref{EPI_faceline}(b) and (c), EPIs correspond to the 3D facial curves from the ground truth. Different line slopes in the EPI indicate different curve shapes. We use 14,000 synthetic light field images corresponding to the $80$ identities of BU-3DFE for training. All together we extract 11.2 Million horizontal and vertical EPIs as training samples. Figure~\ref{EPI_example} shows some example EPIs and their corresponding curves. Using EPI images as training data removes the need for a huge number of identities. Since each 3D face curve can be learned independently from its corresponding EPI, we are able to generate massive training data from a small number of 3D face scans. Note that we do not need any further data augmentation such as image inversion or multiple crops as our networks learn from the full EPIs.

\subsection{FaceLFnet Architecture}
\vspace{-1mm}
Each EPI in our case corresponds to a 3D face curve as shown in Figure~\ref{EPI_faceline} and \ref{EPI_example}. The goal is to predict the full 3D curve from the EPIs  using deep learning. CNNs can learn slope information of the pixels from individual EPIs, however, pixel wise prediction is very challenging. Heber \etal~\cite{Heber2016_CVPR} divided each EPI into patches for 3D scene estimation. The authors estimated the depth value from each EPI patch independently as it contained the information pertaining to a single line at the center of the patch. In our case, pixel wise estimation is not practical as our network must learn the inter-relationship between the lines in one full EPI to estimate the complete 3D curve. Furthermore, in case of light field images for faces, some EPI patches especially in the quasi planar facial areas are devoid of lines and hence do not contain enough depth information leading to inaccurate depth estimation. Therefore, we propose using a complete EPI for depth prediction in order to exploit the correlations of adjacent pixels and mitigate the problem of inaccurate depth estimation due to pixel wise prediction.

\begin{table}[b!]
\small
\begin{center}
\begin{tabular}{c|c|c}
\hline
Layers                              & Output Size & FaceLFnet                                                                                             \\ \hline
Convolution                         & $15\times400$      & $3\times3$ conv, stride 1                                                                                  \\ \hline
Dense Block 1                       & $15\times400$      & {[}$3\times3$ conv, stride 1{]}$\times3$                                                                         \\ \hline
\multirow{2}{*}{Transition Layer 1} & $15\times400$      & $3\times3$ conv, stride 1                                                                                  \\ \cline{2-3}
                                    & $8\times200$       & $2\times2$ average pool, stride 2                                                                          \\ \hline
Dense Block 2                       & $8\times200$       & {[}$3\times3$ conv, stride 1{]}$\times3$                                                                          \\ \hline
\multirow{2}{*}{Transition Layer 2} & $8\times200$       & $3\times3$ conv, stride 1                                                                                  \\ \cline{2-3}
                                    & $4\times100$       & $2\times2$ average pool, stride 2                                                                          \\ \hline
Dense Block 3                       & $4\times100$       & {[}$3\times3$ conv, stride 1{]}$\times3$                                                                          \\ \hline
\multirow{2}{*}{Transition Layer 3} & $4\times100$       & $3\times3$ conv, stride 1                                                                                  \\ \cline{2-3}
                                    & $2\times50$        & $2\times2$ average pool, stride 2                                                                          \\ \hline
Dense Block 4                       & $2\times50$        & {[}$3\times3$ conv, stride 1{]}$\times3$                                                                          \\ \hline
Regression Layer                    & 400         & \begin{tabular}[c]{@{}c@{}}4096 fully-connected\\ 400 fully-connected\\ EuclideanLoss\end{tabular} \\ \hline
\end{tabular}
\end{center}
\caption{Our proposed FaceLFnet architecture. Note that each convolutional layer in the dense block corresponds to the sequence BN-ReLU. The growth rate of the four blocks is $k=12$.}
\label{Table_CNN}
\end{table}

The dimensions of each input EPI are $15\times400\times3$ (horizontal/vertical sub-aperture images $\times$ horizontal/vertical image pixels $\times$ RGB channels). Such a low resolution in the first dimension and size disparity in the first two dimensions pose challenges as the information of the input EPIs will reduce rapidly in one dimension than the other when passed through a deep network. To extenuate this problem and inspired by the success of Gao \etal~\cite{DenseNet2017_CVPR}, we propose a light field face network for estimating facial geometry from EPIs. The architecture of our network is illustrated in Figure~\ref{LFface_CNN}. It is based on DenseNet that consists of multiple dense blocks and transition layers. We use four dense blocks and change the softmax classifier to a regressor. Before passing the EPIs through the first dense block, a $16$ channels convolution layer with $3\times3$ kernel size is used. For each dense block, we use three convolutional layers and set the growth rate to $12$. We also use convolution followed by average pooling as transition layers between two adjacent dense blocks. The sizes of feature-map in the four dense blocks are $15\times400$, $8\times200$, $4\times100$ and $2\times50$ respectively. The details of network configurations are given in Table~\ref{Table_CNN}.

Both horizontal and vertical FaceLFnets are trained from scratch using the Caffe deep learning framework~\cite{Caffe2014}. The initial learning rate is set to 0.0003 which is divided by 10 at 30000 and 50000 iterations. Our networks require only one epoch for convergence. The caffe model for the trained networks will be made public.

\begin{figure}[t]  
  \center{\includegraphics[width=0.45\textwidth]{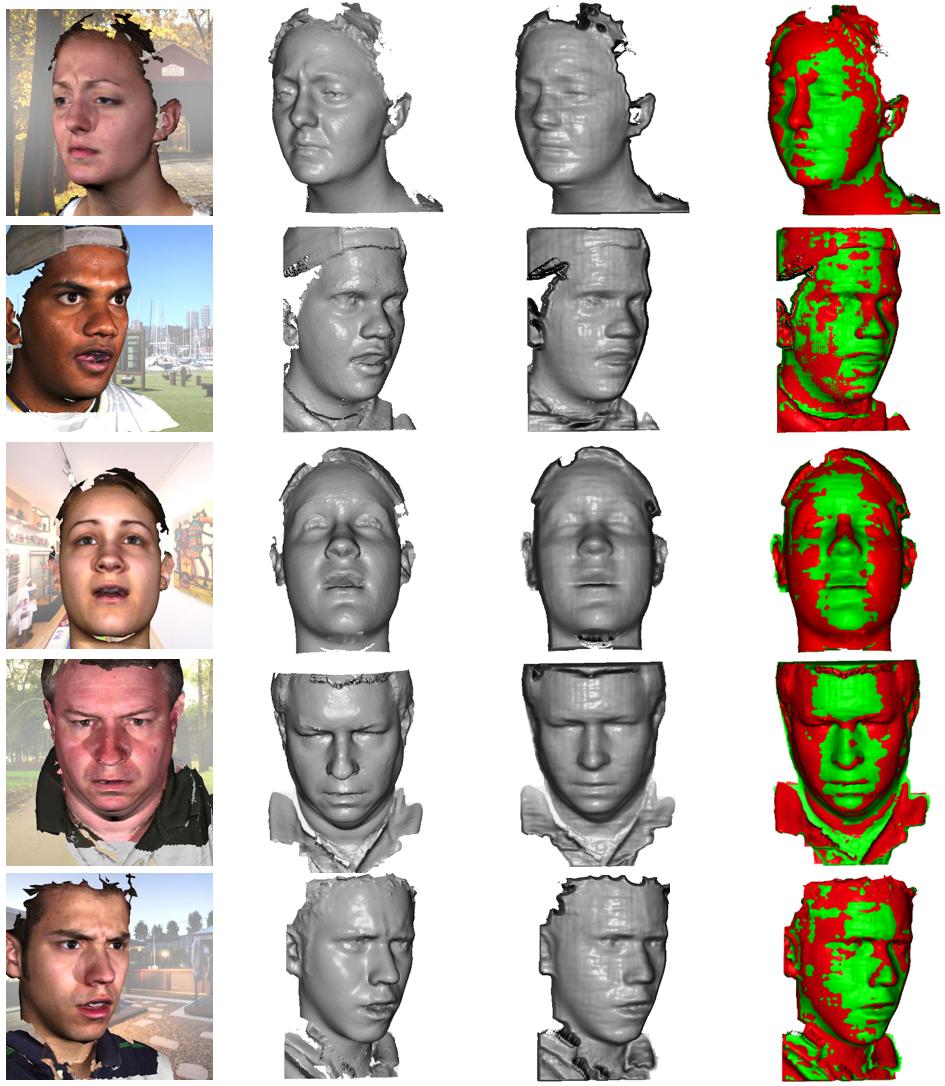}}
  \caption{\label{Pose_invariance} Pose invariance.  Columns one to four in each row respectively depict the input central view of the light field image, the ground truth 3D face, the reconstructed 3D face by our proposed method and the last two overlaid on each other.}
  \vspace{-2mm}
\end{figure}

\begin{figure}  
  \center{\includegraphics[width=0.45\textwidth]{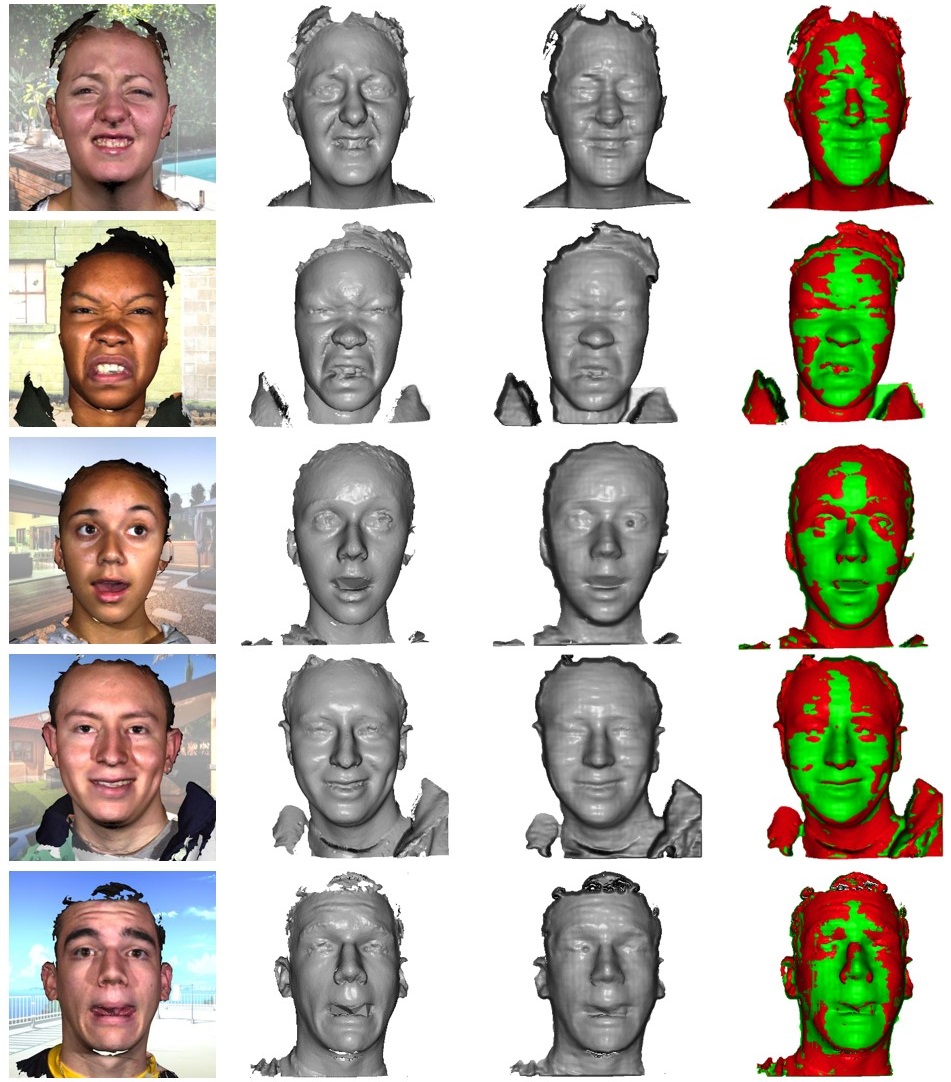}}
  \caption{\label{Expression_invariance} Expression invariance. As shown, our method can handle exaggerated expressions. Columns one to four in each row respectively depict the input central view of the light field image, the ground truth 3D face, the reconstructed 3D face by our proposed method and the last two overlaid on each other.}
\end{figure}

\begin{figure}  
  \center{\includegraphics[width=0.45\textwidth]{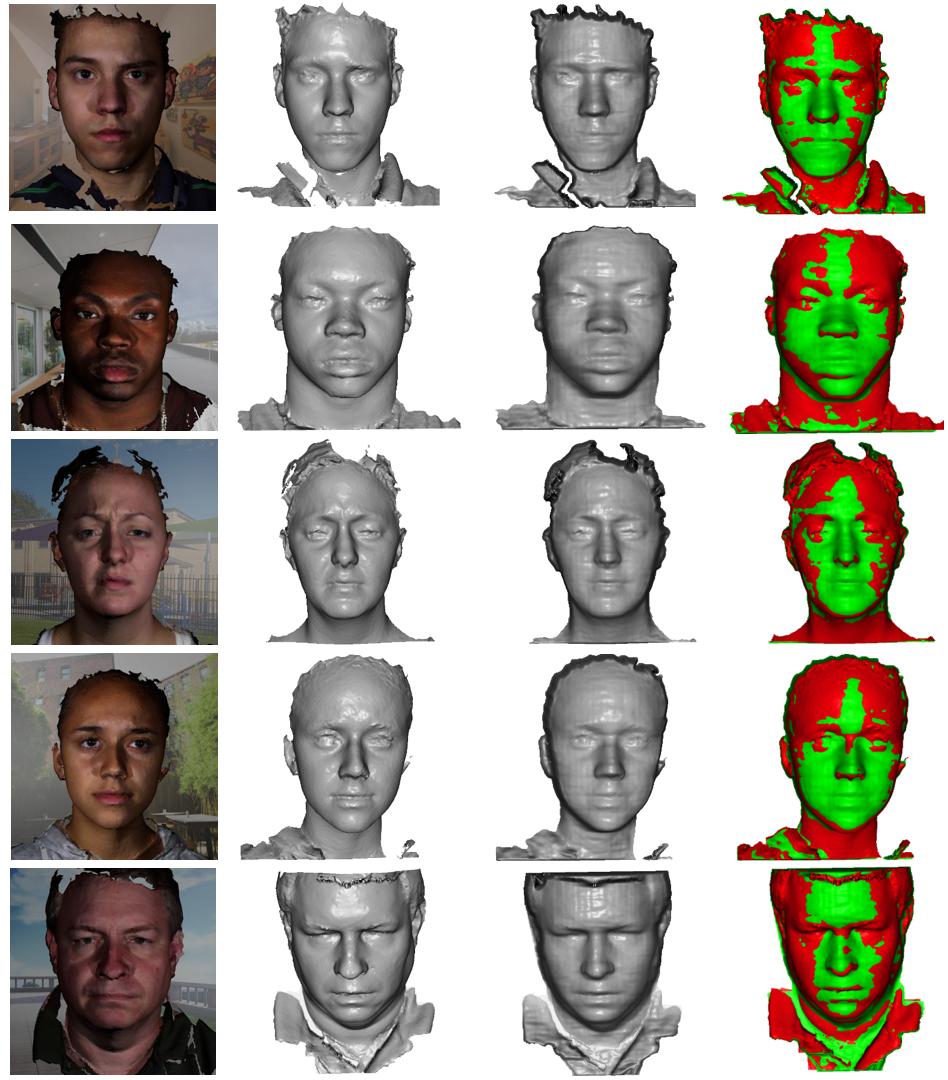}}
  \caption{\label{Light_invariance} Invariance to illumination and skin color. Our method is robust to illumination variations and also works well in the case of dark skin (second row).}
  \vspace{-5mm}
\end{figure}

\subsection{3D Face Reconstruction}
\vspace{-1mm}
The output of our horizontal and vertical FaceLFnets are 3D facial curves that together make up a 3D face. We combine all the horizontal and vertical curves (in our case $400$ each) of a face to form a horizontal and a vertical depth map separately. The next step is to reconstruct a 3D face from the two depth maps. A naive way to reconstruct the face is to take the average of both depth maps. However, such a methodology results in reconstruction error as each curve was learned independently. To mitigate this problem we propose a technique to project the depth maps on a 2D surface. First of all, we convert both the depth maps to 3D pointclouds, using the camera parameters. Next we give a slight jitter to the horizontal pointcloud by translating it $1mm$ to the left on x-axis only. We fit a single surface of the form $z(x,y)$ to both 3D pointclouds simultaneously using the \textit{gridfit} algorithm~\cite{john2008}. Our method ensures that a smooth surface is fitted to the horizontal and vertical pointclouds taking into account the correlation between the curves resulting in a smooth reconstructed 3D face.

\begin{figure*}[t] 
  \center{\includegraphics[width=\textwidth]{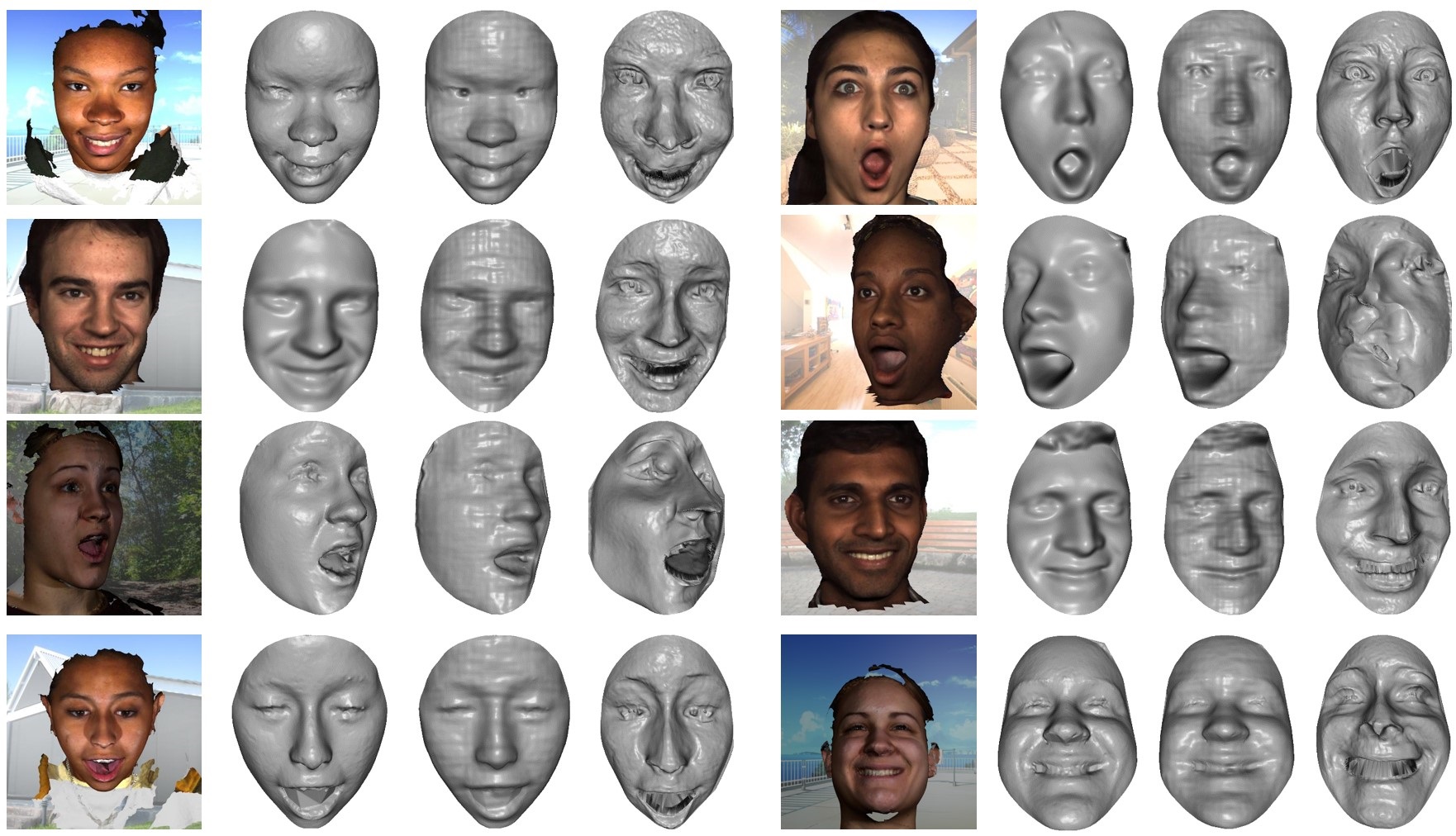}}
   \caption{\label{Evaluation} Qualitative results. The columns contain (in order) central view image, the ground truth 3D face, 3D face reconstructed by our method and 3D face reconstructed by Sela \etal~\cite{FacialI2I2017_ICCV}.}
   \vspace{-2mm}
\end{figure*}

\vspace{-2mm}
\section{Experimental Results}
\vspace{-2mm}
We now present the evaluation of our method for 3D face reconstruction from a single light field image on the 3D scans of the remaining $20$ subjects from BU-3DFE and all $101$ subjects from the BU-4DFE dataset. We compare our subjective results with the recent state-of-the-art algorithm~\cite{FacialI2I2017_ICCV} for qualitative evaluation. We also present quantitative comparison with VRN-Guided~\cite{LP3D2017_ICCV} and other state-of-the-art methods~\cite{FacialI2I2017_ICCV,DetailedFace2017_CVPR,46X2015_eval,Reference3D2011_PAMI,3DDFA2016_eval,EOS2016_eval} on both datasets. Note that the VRN-Guided method incorporates facial landmarks in their proposed VRN architecture whereas we follow a marker-less strategy.

\vspace{-2mm}
\subsection{Qualitative Evaluation}
\vspace{-1mm}

For qualitative evaluation, we show our reconstruction results on light field images synthesized from BU-3DFE~\cite{BU3D_face2006} and BU-4DFE~\cite{BU4D2013} databases. We also show the ground truth and predicted 3D face shapes overlaid on each other using the Scanalyze software. Figure~\ref{Pose_invariance} shows the reconstructed 3D faces under different poses to demonstrate that our method is robust to pose variations. Unlike model based algorithms for 3D face reconstruction \cite{E2E3Dface2017_CVPR,FacialI2I2017_ICCV} from a single RGB image, our method can recover the 3D model of the full head including the peripheral regions such as hair and neck and sometimes even part of the clothing. Figure~\ref{Expression_invariance} shows our results under exaggerated expressions while Figure~\ref{Light_invariance} shows our results under illumination changes. Note that our method is robust to variations in pose, expressions and illumination.

We use the code provided by Sela \etal~\cite{FacialI2I2017_ICCV} for qualitative comparison of the reconstructed faces. Figure~\ref{Evaluation} shows 3D faces reconstructed from light field images using our method and 3D faces reconstructed from single central view RGB images using the recent state-of-the-art method proposed by Sela \etal~\cite{FacialI2I2017_ICCV}.
Since~\cite{FacialI2I2017_ICCV} estimate only the facial region, we also crop our reconstructed faces for better visual comparison. As demonstrated, our method produces more visually accurate reconstructions in the global geometry compared to \cite{FacialI2I2017_ICCV}. As compared to methods based on fine-tuning, our method can not capture fine details since we use the output of our network directly without complex post-processing steps. Our proposed method performs better than~\cite{FacialI2I2017_ICCV} because, firstly, \cite{FacialI2I2017_ICCV} relies on a face detector and crops the input RGB image based on the detected coordinates while our method does not need any face detection or cropping. Secondly, \cite{FacialI2I2017_ICCV} synthesized their training data from 3DMM parameters and thus their training images do not have the neck and hair regions etc. When the input images are far from the model space, the global face shape will be unsatisfactory at some key facial regions like mouth, nose and eyes as can be seen in Figure~\ref{Evaluation}. Finally, Sela \etal~\cite{FacialI2I2017_ICCV} use non-rigid registration to fit the 3DMM to the coarse output of the proposed network. The model fitting process deforms the facial shape when the model and the coarse shape estimated by the network are quite different.

\vspace{-3mm}
\subsection{Quantitative Evaluation}
\vspace{-1mm}

\begin{figure}[t]  
	\center{\includegraphics[width=0.47\textwidth]{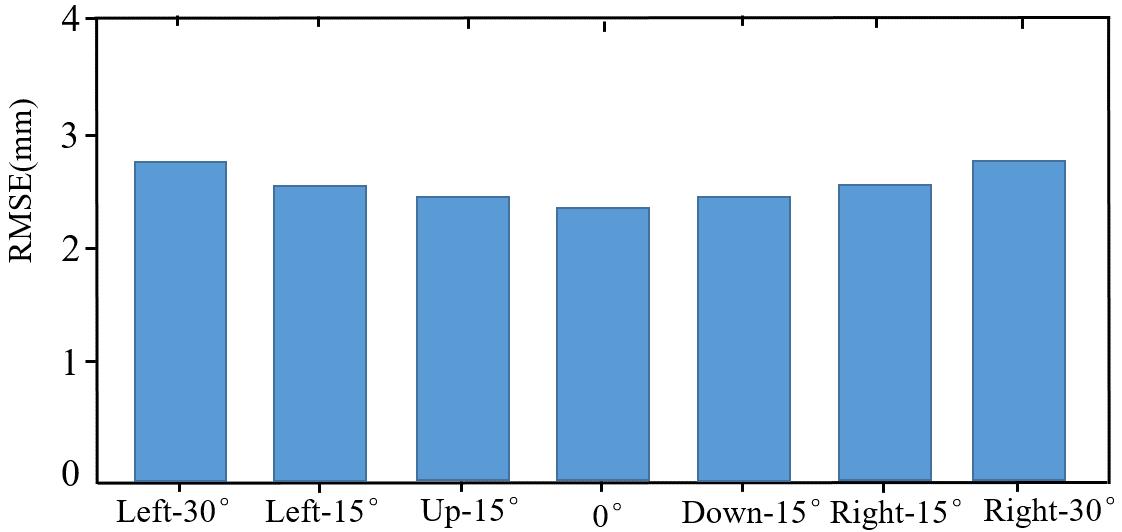}}
	\caption{\label{Pose_metric} Reconstruction errors for different facial poses on the BU-3DFE dataset \cite{BU3D_face2006}. Note that the RMSE increases from 2.62 to 2.93 (by 0.31 mm only) under extreme pose variations.}
\vspace{-2mm}
\end{figure}

For quantitative comparison, we evaluate the 3D reconstruction on 3,500 light field images of $20$ subjects from BU-3DFE~\cite{BU3D_face2006} and 1,400 light field images of $101$ subjects from the BU-4DFE dataset. To measure the affect of pose on the reconstruction accuracy, we use the 3,500 light field images from BU-3DFE dataset. There are 500 light field images for each pose. We use the Root Mean Square Error (RMSE) between the 3D point clouds of the estimated and ground truth reconstructions as a quantitative measure. Results of RMSE for different poses are depicted in Figure~\ref{Pose_metric}. Our method is robust to pose variations as the RMSE error increases by only $0.31mm$ when the pose is varied by 30 degrees.

\begin{figure}[t]  
	\center{\includegraphics[width=0.47\textwidth]{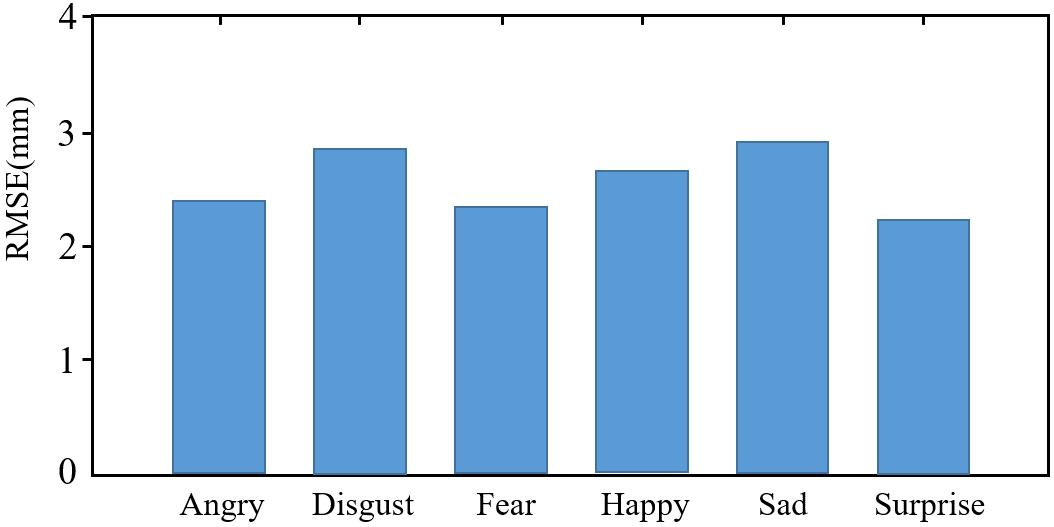}}
	\caption{\label{Expression_metric} Reconstruction errors for different facial expressions on the BU-4DFE dataset \cite{BU4D2013}. The RMSE increases from 2.49 to 2.98 (by only 0.49 mm) under extreme expression variations. Sad has the highest error whereas surprise has the lowest because of more edges around the lips which favors EPI based reconstruction.}
	\vspace{-3mm}
\end{figure}

To measure the affect of facial expressions on reconstruction accuracy, we synthesize frontal images in different expressions (Angry, Disgust, Fear, Happy, Sad and Surprise) from the BU-4DFE dataset and measure the reconstruction errors. Figure~\ref{Expression_metric} shows that the RMSE of 3D face reconstruction from our method is small even in the presence of exaggerated expressions.

We compare the absolute depth error of  our proposed method with the state-of-the-art in Table~\ref{BU3D_evaluation}, which shows that our proposed 3D reconstruction outperforms all existing methods. We report depth errors evaluated by mean, standard deviation, median and the average ninety percent largest error. Note that for a fair comparison with Sela \etal \cite{FacialI2I2017_ICCV} we report the results obtained on the same dataset directly from their paper instead of calculating the reconstruction errors from our implementation of their work. 

We also compare the results of our method with VRN-Guided~\cite{LP3D2017_ICCV}, 3DDFE~\cite{3DDFA2016_eval} and EOS~\cite{EOS2016_eval} methods using the BU-4DFE dataset \cite{BU4D2013}. We use the Normalized Mean Error (NME) metric proposed by Aarson~\cite{LP3D2017_ICCV} to report the results for comparison with existing methods. NME is defined as the average per vertex Euclidean distance between the estimated and the ground truth reconstruction normalized by the outer 3D interocular distance:
\vspace{-2mm}
\begin{equation}\label{NMEfunction2}
\vspace{-2mm}
{\rm NME} =\frac{1}{n}\sum_{k=1}^{n}\frac{\left\| x_{k}-y_{k} \right\|_2}{d},
\vspace{-0mm}
\end{equation}
where $n$ is the total number of vertices per facial mesh and $d$ is the interocular distance. $x_{k}$ and $y_{k}$ represent the coordinates of vertices from the estimated and ground truth meshes respectively. The NME is calculated on the face region only. As shown in Table~\ref{BU4D_evaluation}, our method outperforms the state-of-the-art.

\begin{table}
	\centering
	\scriptsize
	\setlength{\tabcolsep}{4.5pt}
	\renewcommand{\arraystretch}{1.5}
	\begin{tabular}{|l|c|c|c|c|}
		\hline
		\multirow{2}{*}{} & \multicolumn{4}{c|}{Error in mm}        \\ \cline{2-5}
		& Mean & SD & Median & 90\% largest \\ \hline
		Kemelmacher \etal\cite{Reference3D2011_PAMI}     & 3.89      & 4.14     & 2.94        & 7.34    \\ \hline
		Zhu \etal\cite{46X2015_eval}     & 3.85      & 3.23     & 2.93        & 7.91     \\ \hline
		Richardson \etal\cite{DetailedFace2017_CVPR}     & 3.61      & 2.99     & 2.72      & 6.82     \\ \hline
		Matan \etal \cite{FacialI2I2017_ICCV}     & 3.51      & 2.69     & 2.65        & 6.59     \\ \hline
		Ours & \textbf{2.78}      & \textbf{2.04}     & \textbf{1.73}        & \textbf{5.30}      \\ \hline
	\end{tabular}
	\vspace{2mm}
	\caption{Comparative results on the BU-3DFE dataset \cite{BU3D_face2006}. The absolute RMSE between ground truth and  predicted shapes evaluated by mean, standard deviation, median and the average ninety percent largest error of the different methods are presented.}
	\label{BU3D_evaluation}
	\vspace{-2mm}
\end{table}

\begin{table}[h]
	\small
	\centering
	\begin{tabular}{|c|c|c|c|c|}
		\hline
		& 3DDFA\cite{3DDFA2016_eval}  & EOS\cite{EOS2016_eval} & VRN-Guided\cite{LP3D2017_ICCV} & Ours \\ \hline
		NME & 5.14                        & 5.33                   & 4.71                           & \textbf{3.72} \\ \hline
	\end{tabular}
	\vspace{1mm}
	\caption{Reconstruction errors on the BU-4DFE dataset \cite{BU4D2013} in terms of NME defined in Eq. (2). ICP has been used to align the reconstructed face to the ground truth similar to \cite{LP3D2017_ICCV}.}
	\label{BU4D_evaluation}
	\vspace{-5mm}
\end{table}

\vspace{-3mm}
\section{Conclusion}
\vspace{-2mm}
We presented a model-free  approach for recovering the 3D facial geometry from a single light field image. We proposed FaceLFnet, a densely connected network architecture that regresses the 3D facial curves over the Epipolar Plane Images. Using a curve by curve reconstruction approach, our method needs only a few training samples and yet generalizes well to unseen faces. We proposed a photo-realistic light field image synthesis method to generate a large-scale EPI dataset from a relatively small number of real facial identities. Our results show that 3D face reconstruction from light field images is more accurate and allows the use of a model-free approach which is robust to changes in pose, facial expressions, ethnicities and illumination. We conclude that light field cameras are a more appropriate choice as a passive sensor for 3D face reconstruction since they enjoy similar advantages to conventional RGB cameras in that they are point and shoot, portable and have low cost. These cameras are especially a better choice for medical applications where higher accuracy and model-free approaches are desirable. We will make our trained networks and dataset public which will become the first photo-realistic light field face dataset with ground truth 3D facial scans.

\section*{Acknowledgments}
This research was supported by ARC grant DP160101458. The Titan Xp used for this research was donated by NVIDIA Corporation.


\end{document}